\pdfoutput=1
\documentclass[11pt]{article}

\usepackage[final]{acl}

\usepackage{times}
\usepackage{latexsym}

\usepackage[T1]{fontenc}
\usepackage[utf8]{inputenc}

\usepackage{microtype}

\usepackage{inconsolata}

\usepackage{graphicx}
\usepackage{float}
\usepackage{subfigure}
\usepackage{multirow}
\usepackage{amsmath}
\title{AI4Reading: Chinese Audiobook Interpretation System Based on Multi-Agent Collaboration}

\definecolor{porject_url_color}{HTML}{36a6d6}
\newcommand{\changeurlcolor}[1]{\hypersetup{urlcolor=#1}}  
\author{Minjiang Huang\textsuperscript{\rm 1},  Jipeng Qiang\textsuperscript{\rm 1}\thanks{~~Corresponding author.}, \textbf{Yi Zhu}\textsuperscript{\rm 1}, \textbf{Chaowei Zhang}\textsuperscript{\rm 1}, \textbf{Xiangyu Zhao}\textsuperscript{\rm 2}, \textbf{Kui Yu}\textsuperscript{\rm 3} \\
        \textsuperscript{\rm 1}Yangzhou University, \textsuperscript{\rm 2} City University of Hong Kong, 
        \textsuperscript{\rm 3} Hefei University of Technology \\ \texttt{mz120231035@stu.yzu.edu.cn},
    \texttt{\{jpqiang, zhuyi, cwzhang\}@yzu.edu.cn},
    \\ \texttt{xy.zhao@cityu.edu.hk},  \texttt{yukui@hfut.edu.cn}\\
    \changeurlcolor{porject_url_color}\url{https://www.ai4reading.top}\\
}

\begin{document}
\maketitle

\begin{abstract}
Audiobook interpretations are attracting increasing attention, as they provide accessible and in-depth analyses of books that offer readers practical insights and intellectual inspiration. However, their manual creation process remains time-consuming and resource-intensive. To address this challenge, we propose AI4Reading, a multi-agent collaboration system leveraging large language models (LLMs) and speech synthesis technology to generate podcast-like audiobook interpretations. The system is designed to meet three key objectives: accurate content preservation, enhanced comprehensibility, and a logical narrative structure. To achieve these goals, we develop a framework composed of 11 specialized agents—including topic analysts, case analysts, editors, a narrator, and proofreaders—that work in concert to explore themes, extract real-world cases, refine content organization, and synthesize natural spoken language. By comparing expert interpretations with our system's output, the results show that although AI4Reading still has a gap in speech generation quality, the generated interpretative scripts are simpler and more accurate. The code of AI4Reading is publicly accessible \footnote{https://github.com/9624219/AI4reading}, with a demonstration video available \footnote{https://youtu.be/XCLAsRI9v2k}.

\end{abstract}

\section{Introduction}

In recent years, interpretative or retold versions of audiobooks have attracted much attention \citep{ccarkit2020evaluation,walsh23_interspeech}. Different from unabridged, abridged, or summarized audiobooks, the story is reimagined or modernized to enhance clarity and accessibility for a specific audience, such as younger listeners or those unfamiliar with the original context. This type of audiobook not only preserves the essential themes and narrative arc but also translates archaic language, cultural references, or complex passages into a more relatable and engaging format. To create an interpretative version, publishers and narrators typically collaborate with skilled editors and sometimes the original authors to carefully reword and restructure the text. This process involves identifying and retaining key plot points and character developments while simplifying or rephrasing sections that may be less accessible to modern audiences, which is time-consuming and limits scalability, as shown in Figure 1(a).

\begin{figure}
  \centering
   \includegraphics[width=\linewidth]{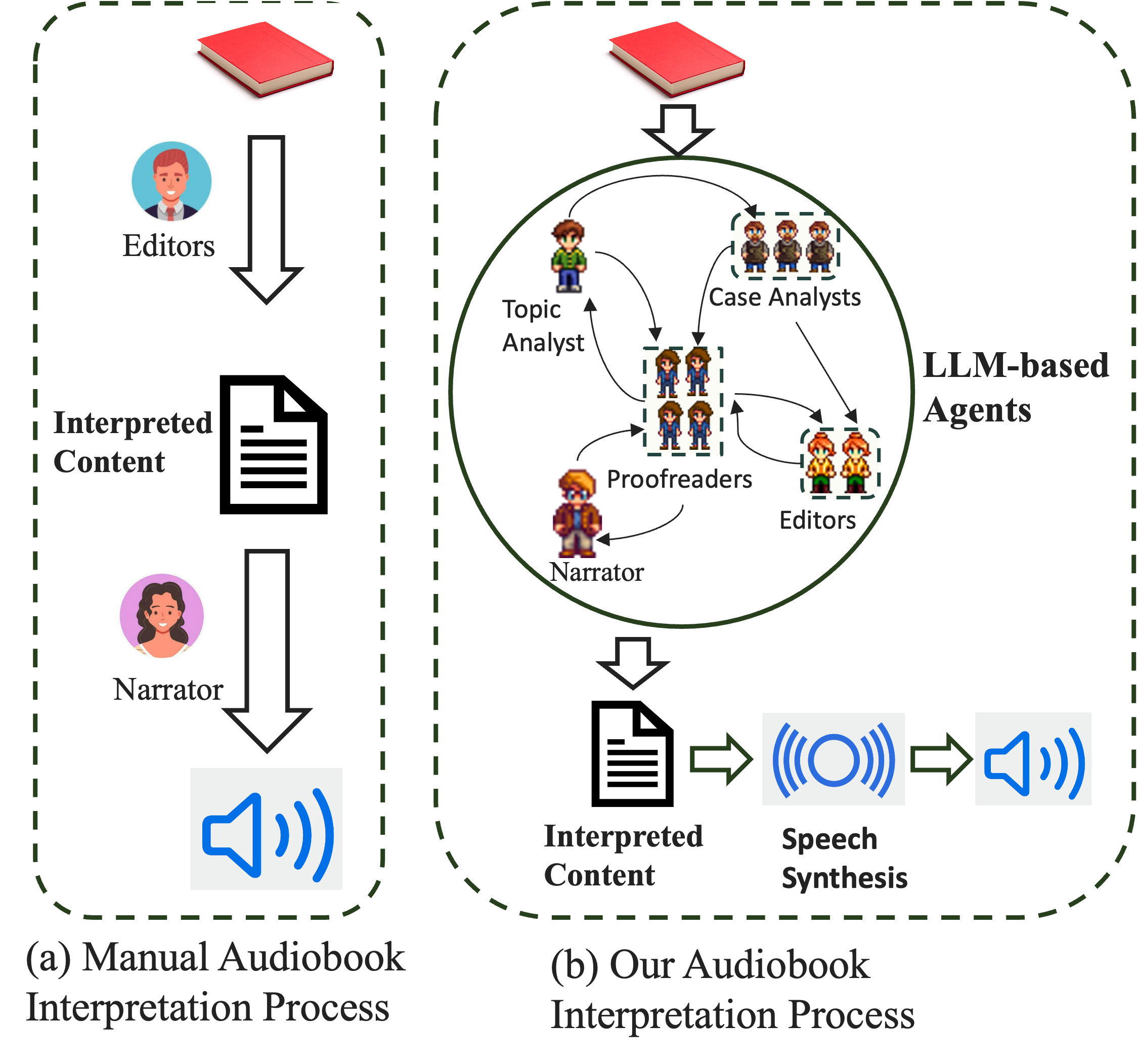} \\
 \caption{Flowchart of expert-based and LLM-based audiobook interpretation system. }
 \label{LLMFig1}
\end{figure}

This paper will explore how to use large language models (LLMs) (such as GPT-4o \citep{achiam2023gpt} or DeepSeek-V3 \citep{liu2024deepseek}) to automatically construct an audiobook interpretation system for these categories of books, including psychology, personal growth, business management, etc. By analyzing experts' interpretations, a good audiobook interpretation system should meet the three key objectives: 

\textbf{(1) Accurate Content Preservation}: It must capture and relay core concepts, theories, and strategies in these fields without oversimplification, ensuring the original insights and depth are maintained. \textbf{(2) Enhanced Comprehensibility}: The system should transform complex ideas into clear, accessible language, enabling listeners to grasp difficult subjects, and provide more practical cases for explanation. \textbf{(3) Logical Narrative Structure}: Maintaining a coherent step-by-step narrative is crucial. This means presenting information in a clear, sequential order that highlights cause-and-effect relationships, so listeners can easily follow the progression of ideas.

Although LLMs have demonstrated strong reasoning capabilities, our tests show that LLMs cannot achieve the above three objectives through chain-of-thought \cite{wei2022chain} or retrieval-augmented generation \cite{jiang2023active} strategies. Multi-agent systems based on LLMs have gradually risen, showing considerable potential for solving complex problems. They have achieved promising results in fields such as software development \citep{hong2023metagpt,nguyen2024agilecoder}, gaming \citep{hua2024game,isaza2024prompt}, and writing \citep{xi2025omnithink,bai2024longwriter}. Therefore, we will design an audiobook interpretation system based on multi-agent collaboration. 

To generate better interpretation manuscripts, we have constructed a combination of 11 agents as shown in Figure 1(b), including: \textbf{Topic Analyst} explores book themes, and provides supporting arguments; \textbf{Three Case Analysts} expand related knowledge, identify and analyze real-world cases to strengthen the core arguments; \textbf{Two Editors} organize content, ensuring logical coherence, clarity, and conversational appropriateness; \textbf{Narrator} converts written content into natural spoken language for an improved listening experience; \textbf{Four Proofreaders} review and ensure accuracy, logical consistency, and adherence to conversational style.

Finally, our contributions are as follows:

(1) We are the first to study how to automatically construct an audiobook interpretation system using large language models and speech synthesis technology. Compared to manual interpretation, this system, AI4Reading, is not only time- and labor-efficient but also overcomes language barriers, enabling the interpretation of books from different languages into the target language. In terms of system capabilities, our approach provides interpretations of both Chinese books and Chinese interpretations of English books.

(2) For the generation of interpretation manuscripts, we propose a multi-agent collaboration approach. To produce engaging interpretative content, this method considers multiple processes, including topic and case identification, preliminary interpretation, oral rewriting, reconstruction and revision. 

(3) We conducted a manual analysis comparing expert interpretations with our results from two aspects: synthesized speech and interpretation manuscripts. The analysis results show that our method produces interpretation manuscripts that are simpler and more accurate. However, the naturalness and appeal of the generated speech are slightly inferior.

\section{Related Work}

\subsection{Audiobook System}
The field of audiobook production has evolved to encompass various narration styles, including unabridged, abridged, summarized, and interpretative (or retold) versions. 

Traditional audiobooks predominantly focus on unabridged and abridged audiobooks, where unabridged versions deliver the full text as written by the author, and abridged versions condense the narrative to reduce listening time while preserving core themes \cite{berglund2021audiobook}.  For example, there are tens of thousands of unabridged audiobooks available on Audible \footnote{\url{https://www.audible.com/}} and Ximalaya \footnote{\url{https://www.ximalaya.com/}} in Chinese.  
Summarized audiobooks, which distill key ideas and insights into concise formats, have also gained traction, particularly for professional and academic contexts. Blinkist\footnote{\url{https://www.blinkist.com/}} is one of the more popular websites in this category. 

More recently, interpretative or retold versions have emerged as a distinct category, wherein the narrative is not merely shortened but is reimagined or modernized to enhance clarity and accessibility for specific audiences \cite{walsh23_interspeech}. This process involves creative editorial adaptations—translating archaic language and complex cultural references into a format that is engaging and relatable, while striving to preserve the original work’s essential themes. FanDeng\footnote{\url{https://www.fanshu.cn/}} platform in Chinese provides such audiobooks, primarily narrated by well-known hosts.

Creating interpretative or retold versions of audiobooks is often the most challenging among the formats discussed. It often necessitates collaboration among authors, editors, and narrators to ensure the adapted version maintains the original's essence while resonating with contemporary listeners. In this paper, we will use a multi-agent approach based on LLMs to automatically generate interpretation manuscripts without human involvement. 

\subsection{Interpretation Generation}

Research related to interpretation content generation includes document summarization \cite{ryu-etal-2024-multi,ravaut-etal-2024-context} and document simplification \cite{fang-etal-2025-collaborative,fang2025progressive,qiang-etal-2023-chinese}. In summarization, the goal is to condense content by selecting key points to produce a shorter version of the original text, and in simplification, the objective is to focus on reducing syntactic and lexical complexity to aid readers with varying language proficiencies or cognitive needs \cite{qiang2023chinese,qiang-etal-2023-parals}. 

Interpretative generation in this paper requires a creative transformation of the original work: it must reimagine and modernize the narrative to suit a target audience while preserving the essential themes, narrative structure, and nuanced details. This process involves not only removing or rephrasing less essential content but also adding clarifications, restructuring passages, and sometimes even introducing new examples to ensure that the story remains engaging and logically coherent. Such a multifaceted task demands a higher level of domain understanding, creative rewriting, and iterative refinement compared to the relatively straightforward tasks of summarization or simplification.

\section{System Design}
\label{system}
\begin{figure*}
  \centering
   \includegraphics[width=\linewidth]{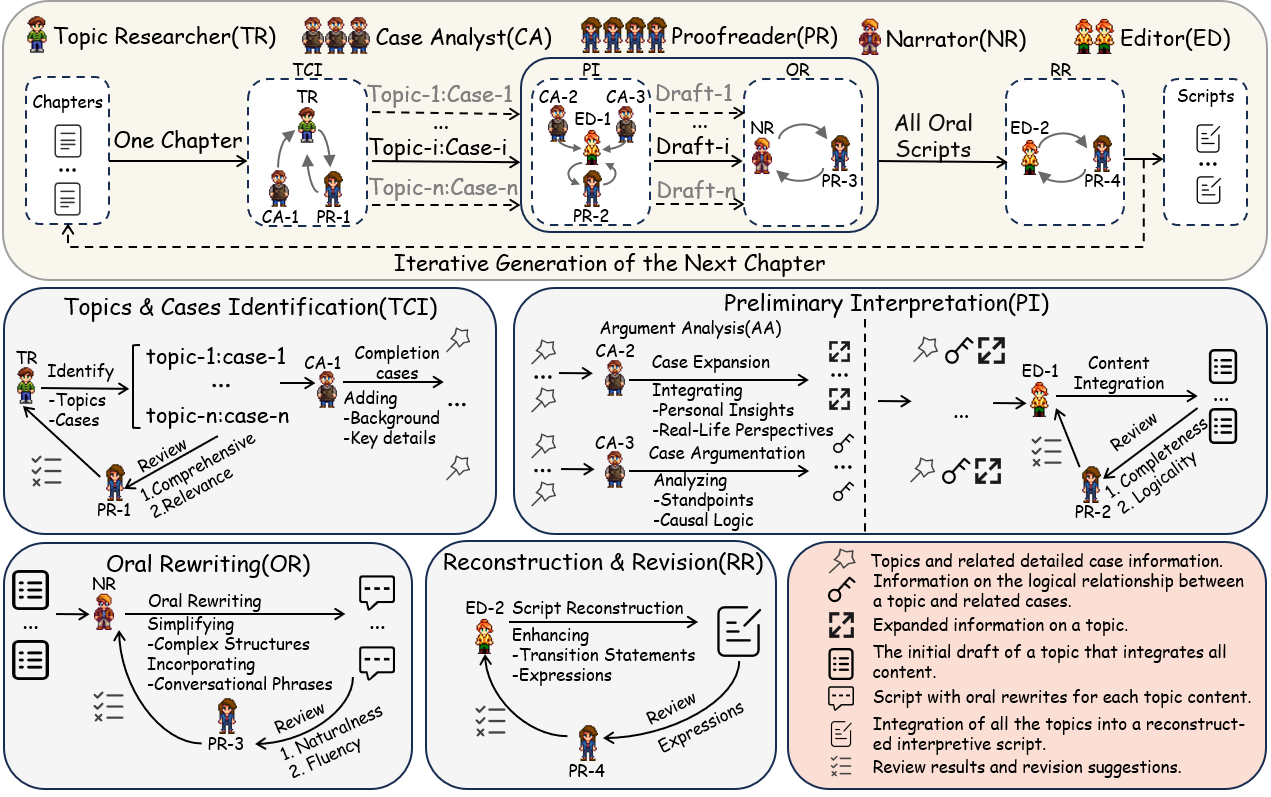} \\
 \caption{The process of interpretation script generation in AI4Reading based on multi-agent collaboration. }
 \label{fig:Agent}
\end{figure*}

This section introduces AI4Reading, an intelligent framework for generating interpretive scripts and audio outputs, capable of automatically transforming book content into structured, naturally expressed interpretive scripts and further producing high-quality audio outputs. The system comprises two core modules: 

(1) \textbf{Interpretation Script Generation:} This module employs a multi-agent collaborative mechanism where specialized roles—such as one Topic Analyst (TA, \raisebox{-0.3ex}{\includegraphics[height=1em]{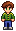}}), three Case Analysts (from CA1 to CA3, \raisebox{-0.3ex}{\includegraphics[height=1em]{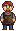}}), four Proofreaders (from PR1 to PR4, \raisebox{-0.3ex}{\includegraphics[height=1em]{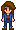}}), one Narrator (NR, \raisebox{-0.3ex}{\includegraphics[height=1em]{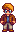}}), and two Editors (ED1 and ED2, \raisebox{-0.3ex}{\includegraphics[height=1em]{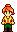}}) —work together to automatically generate the interpretation script.

(2) \textbf{Audio Generation:} This module converts the generated interpretive manuscripts into natural, fluent audio outputs by leveraging Text-to-Speech (TTS) technology. 

\subsection{Interpretation Script Generation}

We propose a collaborative multi-agent framework for generating interpretive scripts, as illustrated in Figure \ref{fig:Agent}. This framework takes chapter content as input and leverages specialized system prompts to assign distinct roles and responsibilities to each agent. A detailed description of each stage is presented below.

\subsubsection{Topics \& Cases Identification (TCI)}

This stage mimics human cognitive processes of reading and summarization, distilling core topics and associated supporting cases, which is carried out by three agents: TA, PR-1, and CA-1. 

TA processes one chapter $S$ of one book to identify a set of core topics $T$ and a preliminary set of relevant cases $C$, which is modeled as: $Agent_{TA}(S) \rightarrow (T, C)$, where $T=\{ t_1, t_2, \ldots, t_n \}$ is the set of core topics extracted from $S$, $C =\{ c_1, c_2, \ldots, c_n \}$ is the set of preliminary cases associated with the topics, $n$ is the number of extracted topics, with a maximum of 3.

To review whether there are unreasonable topic-case pairs in $(T, C)$, we define an agent Proofreader (PR-1) who rigorously reviews each topic-case pair $(t_i, c_i) \in (T, C)$ in terms of comprehensiveness and relevance. This validation process is defined as: 
$Agent_{PR-1}(T, C) \rightarrow \{(t, c)\}_{\text{val}} \cup \{(t, c)\}_{\text{inv}}$, where "$\{(t, c)\}_{\text{val}}$" and "$\{(t, c)\}_{\text{inv}}$" denote the valid and invalid topic-case pairs, respectively. If set "$\{(t, c)\}_{\text{inv}}$" is not empty, TA will be called again. 

While TA and PR-1 ensure topical relevance and comprehensiveness, preliminary cases often lack depth or context. To fill these informational gaps, we define an agent Case Analyst (CA-1) who enriches the cases with additional background information and key details: $Agent_{CA-1}(S,T,C) \rightarrow (T,C')$. CA-1 ensures that the final output aligns with the original content while effectively supporting subsequent tasks.

\subsubsection{Preliminary Interpretation (PI)}

The output $(T,C')$ from the previous stage did not consider how to better facilitate the understanding of theoretical content. In this stage, we aim to supplement each "topic-case" pair by incorporating personal anecdotes and real-life scenarios using these agents (CA-2, CA-3, ED-1, and PR-2), making the content more relatable to everyday life. 

CA-2 expands upon each topic-case $(t_i, c'_i)$ by integrating personal insights and real-life perspectives: $Agent_{CA-2}(t_i, c'_i) \rightarrow a_i$, where $a_i$ represents the expansion for case $c'_i$. CA-3 constructs logical arguments to demonstrate how each case supports its corresponding topic. This involves analyzing standpoints, causal logic, and ensuring consistency with the chapter content and topics: $Agent_{CA-3}(t_i, c'_i) \rightarrow l_i$, where $l_i$ represents the logical argument established for topic $t_i$. 

Considering that $(a_i,l_i)$ lacks the continuity and narrative flow required for a cohesive interpretive manuscript, we define a new Editor, ED-1, who synthesizes all prior analytical findings into a coherent and well-structured preliminary draft for each topic. $Agent_{ED-1}(t_i,c'_i,l_i, a_i) \rightarrow d_i$. where $d_i$ is the preliminary draft of topic $t_i$.

To further ensure rigor and clarity, PR-2 evaluates each topic draft $d_i$ based on two dimensions: completeness and logicality. For drafts that do not meet the required standards, PR-2 provides constructive feedback: $Agent_{PR-2}(d_i) \rightarrow f_i$, where $f_i = (compt_i, log_i, sr_i)$, $compt_i$ and $log_i$ belonging to $\{ "Yes", "No" \}$ indicate whether the draft satisfies the completeness and logicality criterion, $sr_i$ contains specific revision suggestions only if $compt_i = "No"$ or $log_i = "No"$, otherwise, $sr_i = \emptyset$.

Based on the feedback $f_i$, ED-1 iteratively refines the draft $d_i$ to be improved through multiple rounds of optimization:
\begin{equation}
d_i=\left\{
\begin{aligned}
d_i & & \text{if } sr_i = \emptyset, \\
Agent_{ED-1}(d_i, sr_i) & & \text{if } sr_i \neq \emptyset .
\end{aligned}
\right.
\end{equation}

The optimization process continues until $d_i$ passes review ($ sr_i = \emptyset$) or reaches the maximum number of allowable iterations $I_{\max}$. 

\subsubsection{Oral Rewriting (OR)}

In this stage, two agents, NR and PR-3, refine the draft $d_i$ to make its expression more conversational and easier for the audience to understand.

NR performs a conversational paraphrase of $d_i$ by: (1) simplifying complex structures, and (2) integrating colloquial lexicon and conversational markers, which is modeled as: $Agent_{NR}(d_i) \rightarrow o_i$
where $o_i$ represents the oral script of $d_i$.

To ensure high-quality oral output, PR-3 conducts rigorous evaluation of oral output $o_i$, concentrating on two critical dimensions: linguistic naturalness and delivery fluency: $Agent_{PR-3}(\mathcal{O}) \rightarrow G$, where $g_i$ is the feedback of $o_i$.

Based on the feedback $g_i$, NR iteratively refines the oral script $o_i$. This process continues until one of the following termination conditions is met: (1) PR-3 considers the requirements to be met; (2) The maximum number of allowable iterations $I_{max}$ is reached.

\subsubsection{Reconstruction and Revision (RR)}

Upon completing the oral rewriting of multiple scripts $\mathcal{O} = \{ o_1, o_2, \ldots, o_n \}$, the phase moves into the reconstruction and revision phase for the final interpretive manuscript. This stage involves ED-2 and PR-4.

ED-2 integrates all independent interpretive segments $\mathcal{O} = \{ o_1, o_2, \ldots, o_n \}$ into a coherent and unified full-length document. The process begins by selecting the first segment $o_1$ as the initial draft, and subsequent segments $\{ o_2, \ldots, o_n \}$ are incrementally incorporated into the current manuscript according to predefined integration guidelines: 
\begin{equation}
M_i=\left\{
\begin{aligned}
o_1 & &  i = 1, \\
Agent_{ED-2}(M_{i-1}, o_i) & & i>1 .
\end{aligned}
\right.
\end{equation}
where $M_i$ denotes the manuscript after incorporating the $i$-th segment $o_i$. Each integration step ensures logical clarity and natural transitions, continuing until all segments are included: $M=M_n$. Through this process, ED-2 helps the manuscript maintain a seamless narrative flow while preserving the depth and richness of the content.

To prevent inconsistencies during the integration process, we utilized PR-4 to evaluate the entire manuscript $M$ and provide feedback on overall coherence, fluency, and naturalness. The iterative refinement process follows the same mechanism as in the PI and OR stages. Through this series of adjustments, the final interpretation script will achieve better structural coherence and fluency.

\subsection{Audio Generation}

After the interpretation script is generated, we need a TTS tool to convert the script $M$ into audio. Modern TTS technology not only produces natural and smooth speech but also adjusts the tone and emotional expression according to the content's characteristics, providing listeners with a richer and more vivid auditory experience. In our system, we adopt high-quality Fish-Speech \cite{liao2024fish} as TTS tool. 

Additionally, we add transition sound effects at the beginning and end of each chapter to help listeners more clearly perceive the transitions between chapters, thus improving the overall comfort and logical flow of the listening experience. This design not only enhances the user’s listening experience but also increases the coherence of the content and the efficiency of knowledge absorption.

\subsection{Agent Configuration Rationale}
The selection of 11 specialized agents in our AI4Reading framework was a deliberate design choice, stemming from our goal to emulate the collaborative and iterative workflows of human expert teams involved in creating high-quality interpretations. We aimed to decompose the complex task of generating an audiobook interpretation into more manageable, focused sub-tasks, each addressable by an agent with a specific role and set of responsibilities.

Our initial explorations considered simpler architectures, particularly relying on a single, powerful LLM to generate the entire interpretation for a chapter using comprehensive prompts with chain-of-thought \cite{wei2022chain} strategy. However, this approach proved unsuitable for our specific task of interpretation generation for several critical reasons: (1) Tendency towards Summarization, Lacking Interpretative Depth: We observed that even with explicit instructions to "interpret" and "explain," a single LLM often defaulted to producing a high-quality summary of the chapter. This output, while concise and accurate in terms of content distillation, inherently lacked the crucial characteristics of an interpretation. Interpretations require going beyond mere summarization to include elaborations, real-world examples, connections to broader concepts, and a narrative style designed to enhance listener comprehension and engagement, which aligns with our objectives but is contrary to the nature of a summary. (2) Insufficient Content Volume and Coverage: The generated text from a single LLM pass was frequently too brief to adequately cover the nuances and key arguments presented throughout an entire book chapter. Interpretations, by their nature, often expand on the original text to clarify complex points, thus requiring a more substantial word count than a summary. The single-LLM outputs often felt like condensed overviews rather than thorough, engaging explanations.

The current structure, therefore, addresses these shortcomings. Each agent has a clearly defined, relatively narrow scope, allowing for more precise prompting and more reliable execution of its specific function. This granular approach, with iterative feedback loops provided by Proofreader agents, was found to yield more consistent, structured, and truly interpretative scripts that are richer in content, better cover the source material, and more effectively meet the multifaceted requirements of audiobook interpretation. Future work may explore optimizations to this configuration, but the current setup provides a robust foundation to overcome the limitations of simpler, single-pass approaches.

\section{Experiments and Evaluation}

\begin{table*}[ht!]
\centering
\begin{tabular}{ccccccccc}
\hline
\multicolumn{1}{c}{\multirow{2}{*}{Annotators}} & \multicolumn{1}{c}{\multirow{2}{*}{Methods}} & \multicolumn{3}{c}{Audio Quality} & \multicolumn{4}{c}{Textual Content} \\ \cline{3-9} 
\multicolumn{1}{c}{} & \multicolumn{1}{c}{} & Nat. & Conc. & Compn. & Simp. & Compt. & Acc. & Coh. \\ \hline
\multirow{2}{*}{1} & FD & 5.0 & 4.3 & 2.9 & 4.2 & 3.4 & 4.0 & 4.2 \\
 & OURS & 4.2 & 3.8 & 3.8 & 4.2 & 3.2 & 4.4 & 4.8 \\ \hline
\multirow{2}{*}{2} & FD & 4.7 & 4.3 & 4.0 & 4.0 & 4.0 & 4.2 & 4.0 \\
 & OURS & 3.9 & 3.8 & 3.8 & 4.6 & 4.6 & 4.2 & 4.8 \\ \hline
\multirow{2}{*}{3} & FD & 5.0 & 4.2 & 3.1 & 5.0 & 3.6 & 3.8 & 3.8 \\
 & OURS & 4.6 & 3.6 & 2.6 & 5.0 & 3.6 & 4.0 & 4.0 \\ \hline
\multirow{2}{*}{4} & FD & 4.8 & 3.9 & 2.7 & 5.0 & 4.8 & 4.6 & 4.6 \\
 & OURS & 3.6 & 2.5 & 1.8 & 4.8 & 4.4 & 4.8 & 4.2 \\ \hline
\multirow{2}{*}{5} & FD & 5.0 & 4.1 & 4.0 & 3.6 & 3.2 & 4.2 & 3.8 \\
 & OURS & 4.2 & 3.3 & 3.4 & 4.6 & 4.0 & 4.2 & 4.2 \\ \hline
\multirow{2}{*}{Average} & FD & 4.9 & 4.2 & 3.3 & 4.4 & 3.8 & 4.2 & 4.1 \\
 & OURS & 4.1 & 3.4 & 3.1 & 4.6 & 4.0 & 4.3 & 4.4 \\ \hline
\end{tabular}
\caption{Results of human evaluation on two dimensions: audio quality and interpretation script. "FD" is from \url{https://www.fanshu.cn/}}
\label{tab:human_evaluation_result}
\end{table*}

\subsection{Experimental Setup}

We will evaluate our system manually from the following two aspects: audio quality and interpretation script. 

\textbf{System Configuration:} 
Our system is built upon MetaGPT \cite{hong2023metagpt} with Deepseek-V3 \cite{liu2024deepseek} as the base LLM, with the model temperature set to 1.3, max\_token set to 8192, $I_{max}$ set to 3, and the prompting strategy using 0-shot prompting. All used prompts are open-sourced on GitHub. 

\textbf{Benchmark:} 
The competitive product FanDeng\footnote{https://www.fanshu.cn/} (FD) serves as the benchmark for comparison. FD is China’s premier knowledge service platform, founded by Dr. Deng Fan, a renowned media scholar, TV host, and communication expert. All the compared contents are narrated by Fan Deng himself.

\textbf{Data:} We selected interpretative books from the FD website as evaluation material, including 10 explanatory excerpts randomly sampled from 10 chapters across five books, covering topics such as personal growth and business finance. The average duration of our system-generated segments was 4 minutes 59 seconds, and the FD-provided segments averaged 4 minutes 33 seconds.

\subsection{Evaluation Metrics}
To evaluate the quality of the generated speech and interpretation text, we developed an evaluation system \footnote{http://49.232.199.229:14444/, username:admin1, password:1admin} where users rate the speech and interpretation text without knowing whether the results are from our method or FD. 

We conducted a human evaluation using a 1-5 Likert scale with 7 undergraduate participants (4 male, 3 female) from diverse academic backgrounds (e.g., computer science, engineering, business, etc.). All annotators are Chinese speakers. We recorded the time users spent on each webpage interface in the system backend. Users were unaware of this in advance.

\textbf{Audio Evaluation:} The audio generated by our system and that from FD were presented in a randomized order. Users were asked to listen to the two audio clips sequentially, with the order of presentation also randomized. After listening to each clip, users completed a survey assessing the following three dimensions: (1) Naturalness (Nat.): Evaluates whether the audio sounds natural and fluent. (2) Concentration (Conc.): Assesses whether the user felt fatigued or distracted during the listening process. (3) Comprehension (Compn.): Measures the user’s understanding of the audio content.

\textbf{Interpretation Script Evaluation:} Users were initially required to read the original text of the chapters to ensure a thorough understanding of the source content. Users then responded to questions based on the selected script. The evaluation encompassed the following four dimensions: (1) Simplicity (Simp.): Assesses the effectiveness of the interpretation script in reducing the comprehension difficulty of the original text. (2) Completeness (Compt.): Checks whether the interpretation script omitted any key information from the original chapters, as identified by the evaluators. (3) Accuracy (Acc.): Determines whether the main ideas conveyed by the interpretation script were consistent with those of the original text. (4) Coherence (Coh.): Analyzes whether the interpretation script contains any logical inconsistencies, such as abrupt content shifts, broken causal relationships, or contradictions.

\subsection{Results}

Although there were seven users, we observed that two users had significantly shorter evaluation times, suggesting they may not have completed the tasks conscientiously. Consequently, valid data from only five users were retained for analysis. 

The results are shown in Table \ref{tab:human_evaluation_result}. In the audio evaluation, our system achieved better results in terms of simplification, while other metrics were lower than FD's results. However, it is also evident that our system is a highly effective approach for generating speech. Regarding textual generation, our method outperformed in all four metrics, demonstrating that our multi-agent-based approach is highly effective for generating interpretative scripts. The evaluation results fully demonstrate the advantages of multi-agent collaboration in content creation and validate the effectiveness of our framework.

\section{Conclusions}
In this paper, we propose a novel multi-agent collaborative system for interpretative audiobook generation, addressing the critical challenges of cost, quality, and language accessibility in traditional audiobook production. By simulating the workflow of human-authored interpretation scripts through specialized agents, including Topic Researchers, Case Analysts, Editors, etc. The system achieves efficient and accurate content distillation. 

\section*{Acknowledgement}

This research is partially supported by the National Natural Science Foundation of China (62076217), the National Language Commission of China (ZDI145-71), the Blue Project of Jiangsu province, and the Top-level Talents Support Program of Yangzhou University.

\section*{Limitations}
Our study acknowledges several limitations that must be addressed in future research. First, the evaluation sample was relatively small, which may not fully capture the diversity of listener experiences. A broader validation involving a larger and more varied group of participants is essential to establish the generalizability and robustness of our system. Second, the current method has been tested exclusively on data from psychology, personal growth, and business management books. This domain constraint limits the system’s applicability, as it cannot yet process or generate interpretations for literature or novels. Expanding the system to accommodate these additional genres will be a critical focus of future research efforts.

\section*{Ethical Considerations} 

In developing AI4Reading, we prioritize ethical responsibility in several key areas. First, we ensure that all content generated by the system complies with copyright law and is not used for commercial purposes. Given that our system reinterprets texts, we are committed to maintaining the integrity and core messages of the original works while making them more accessible to listeners. Transparency is another critical factor—we clearly indicate when content is AI-generated to prevent misinformation. Through these measures, we strive to balance innovation with ethical responsibility, fostering trust in AI-driven audiobook interpretation.

\bibliography{main}
\clearpage
\appendix

\section{System Evaluation}

\subsection{Time Spent of User Evaluation Dataset}

\begin{figure}[ht!]
    \centering
    \includegraphics[width=\linewidth]{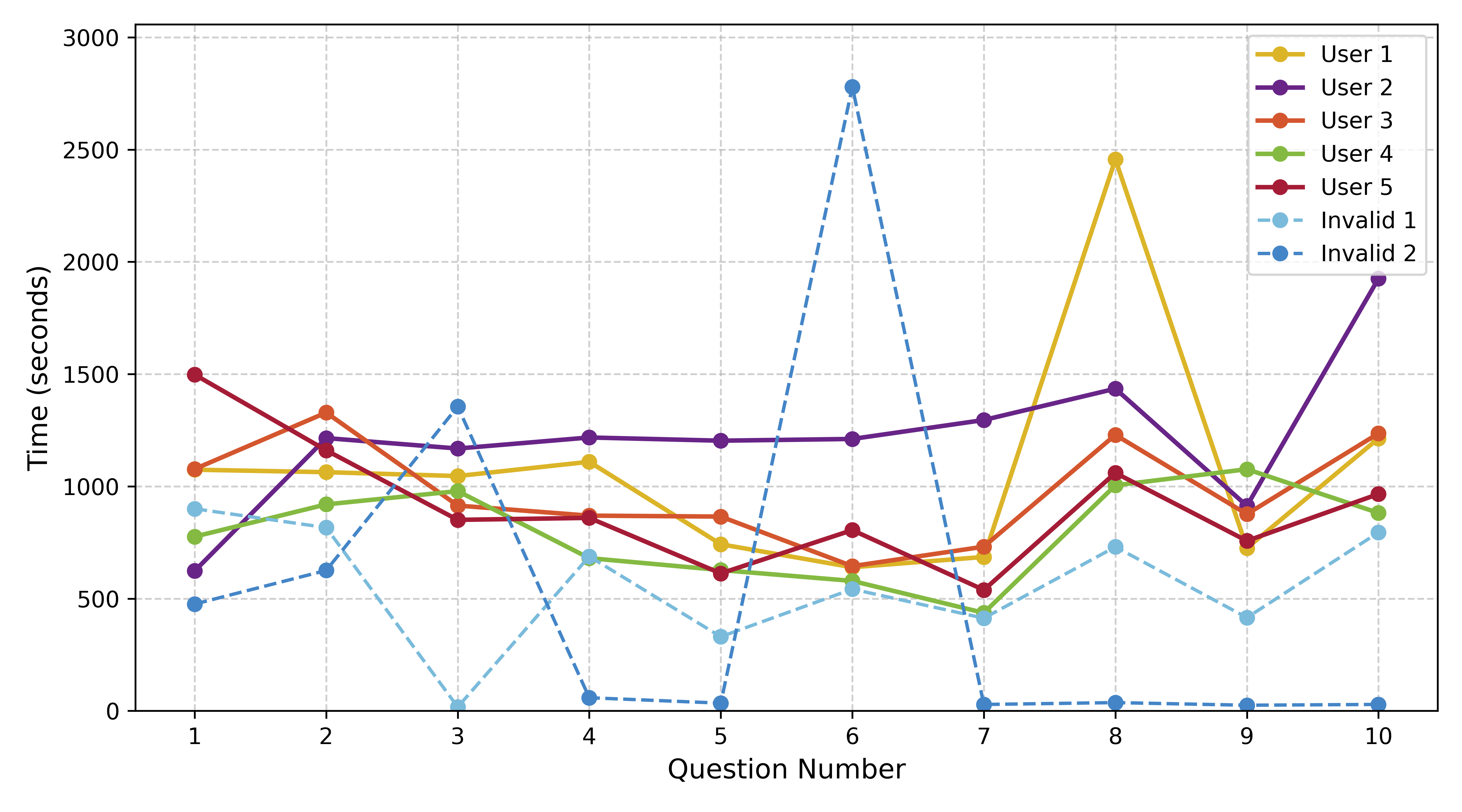}
    \caption{The time spent by the 7 evaluators on each element of the evaluation, with invalid 1 and invalid 2 being the users who discarded the results.}
    \label{fig:time}
\end{figure}

The evaluation time for each user is shown in Figure \ref{fig:time}. The reading time of two users was very short, so they were considered invalid users, and their evaluation results were deemed invalid.

\subsection{System Interface for Audio Evaluation }

\begin{figure}[ht!]
    \centering
    \includegraphics[width=\linewidth]{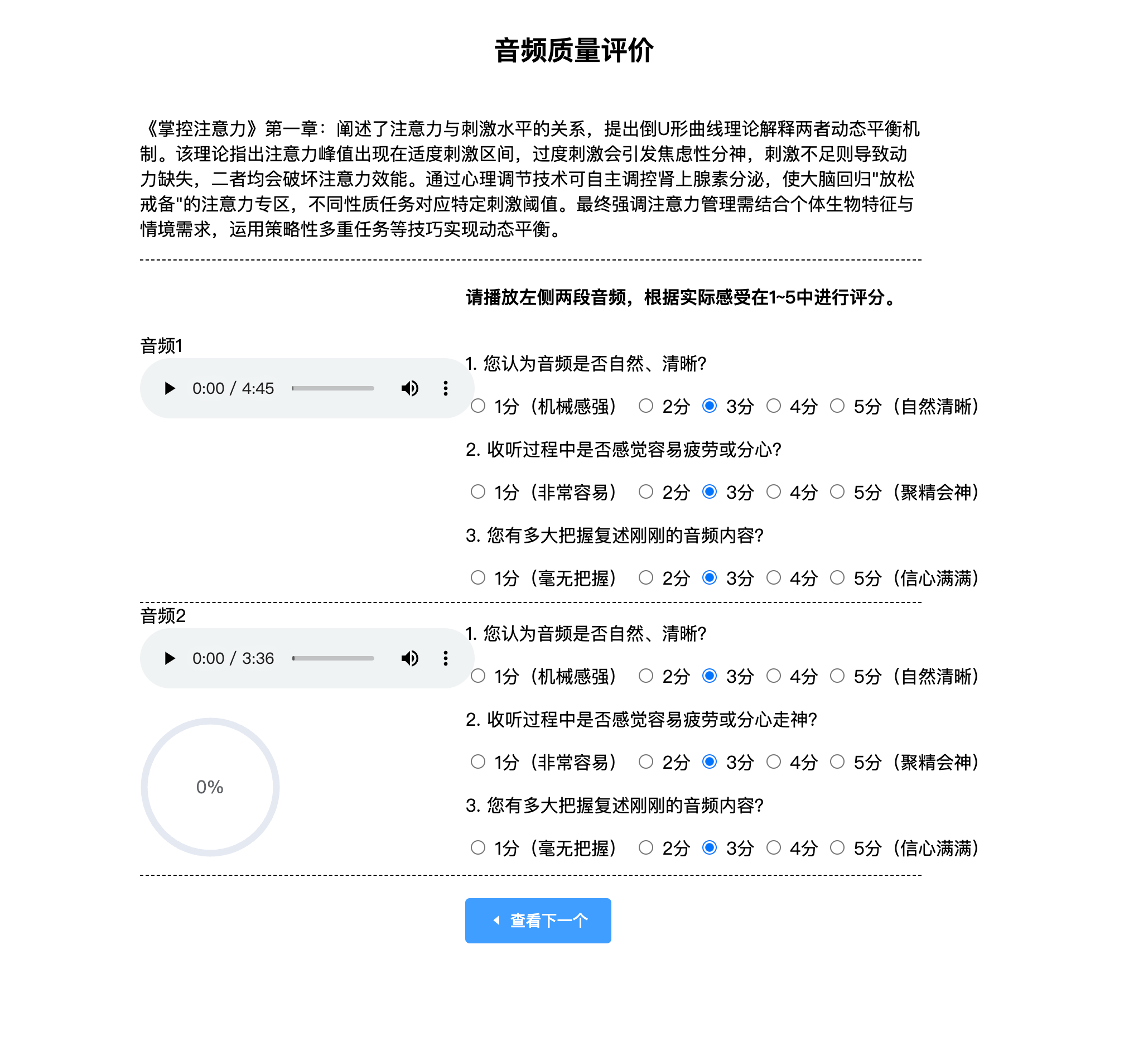}
    \caption{Screenshot of audio evaluation.}
    \label{fig:sae}
\end{figure}

As shown in Figure \ref{fig:sae}, users listened to randomly ordered audio clips from our system and FD, then completed a survey evaluating three aspects: (1) Naturalness—how fluent and natural the audio sounded, (2) Concentration—whether they felt fatigued or distracted, and (3) Comprehension—how well they understood the content.

\subsection{System Interface for Interpretation Script Evaluation }

\begin{figure}[ht!]
    \centering
    \includegraphics[width=0.7\linewidth]{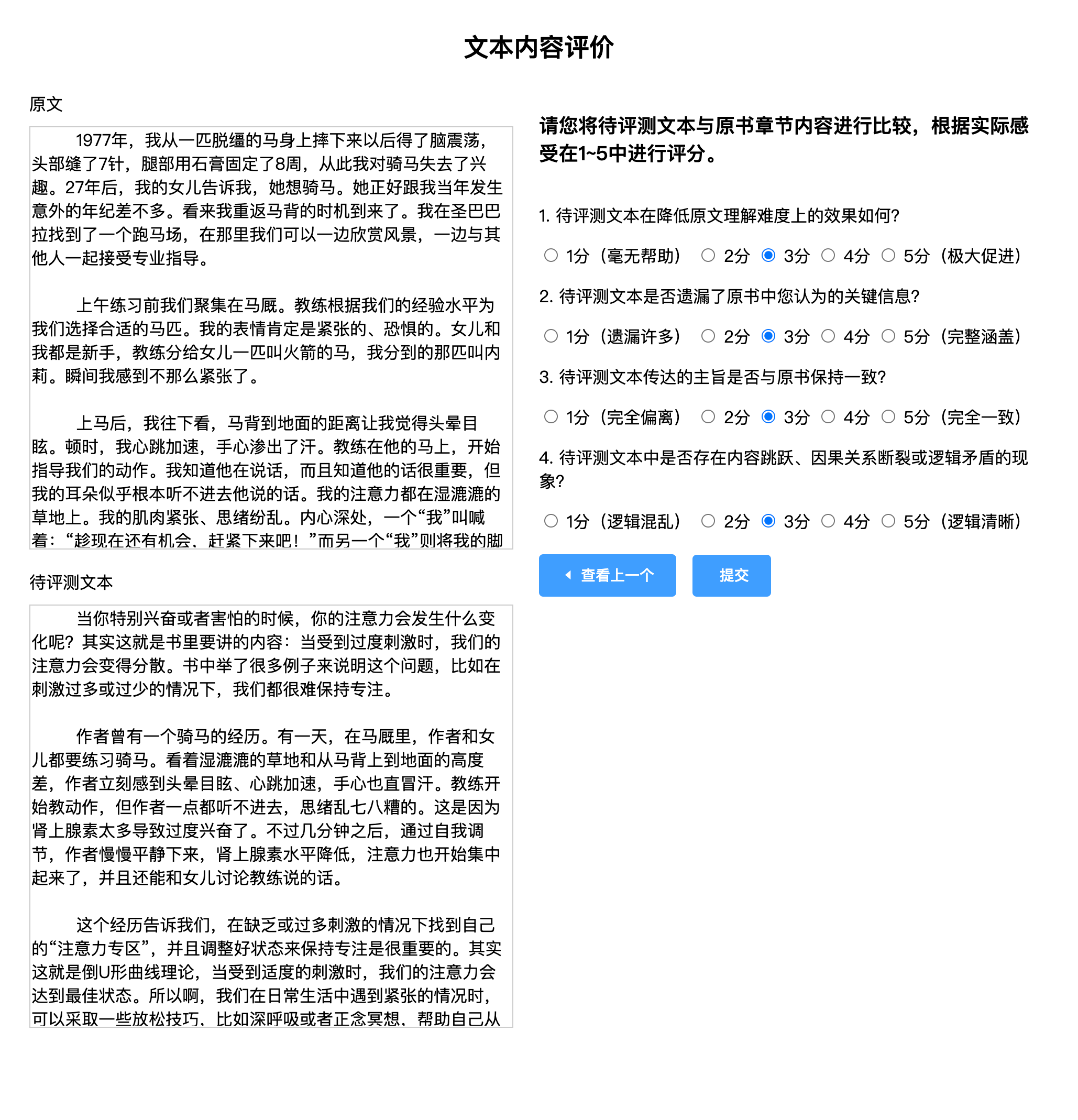}
    \caption{Screenshot of interpretation script evaluation.}
    \label{fig:isa}
\end{figure}

As shown in Figure \ref{fig:isa}, users first read the original chapters to ensure a thorough understanding before evaluating the interpretation script. The assessment covered four dimensions: (1) Simplicity—how effectively the script reduced comprehension difficulty, (2) Completeness—whether key information was omitted, (3) Accuracy—consistency of main ideas with the original text, and (4) Coherence—absence of logical inconsistencies or contradictions.

\section{System Interface}
\label{sec:System Interface}
% \begin{figure}[ht!]
%     \centering
%     \includegraphics[width=\linewidth]{latex/figures/mindmap.png}
%     \caption{Screenshot of the book's mindmap.}
%     \label{fig:Mindmap}
% \end{figure}
\begin{figure*}[ht!]
    \centering
    \includegraphics[width=\linewidth]{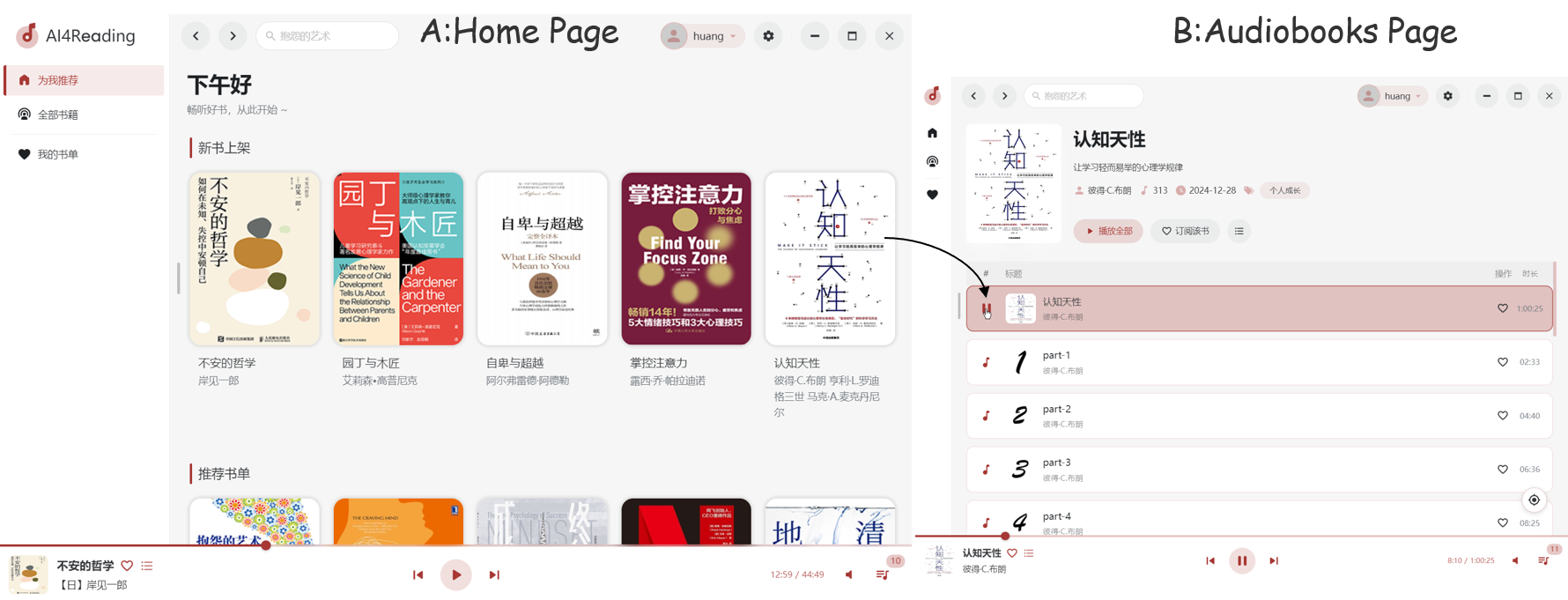}
    \caption{Screenshots of system interface.}
    \label{fig:system_interface}
\end{figure*}

We have designed a concise, intuitive user interface\footnote{If the HTTPS URL is inaccessible, you may try the HTTP URL as an alternative: http://49.232.199.229:14558/ } to optimize the user experience, as illustrated in Figure \ref{fig:system_interface}. The homepage (A) displays a list of books processed by the system. By clicking on a book cover, users can directly access the Audiobooks page (B) to access audio interpretations and related mind maps. This straightforward design enables users to quickly locate their desired books while significantly reducing operational complexity.

On the audiobook page, we offer two interpretation modes: full-book interpretation and chapter-by-chapter interpretation. Users can browse the audio list for a book and listen to the corresponding content by clicking the play button. Each audio entry is clearly labeled with the title, duration, and author information, allowing users to select specific chapters based on their needs. Additionally, the interface includes practical features like bookmarking, sharing, and subscribing to enhance usability and interactivity.

By combining auditory and visual sensory experiences, our design provides high-quality audio interpretations while leveraging mindmap to help users intuitively organize the core content and logical structure of the books. This multimodal learning approach enhances users' understanding of the material, improving both learning efficiency and overall reading experience. Whether for fragmented learning or systematic reading, the interface caters to diverse user needs, providing an immersive learning experience.

% \section{Mind Map Generation}
% Based on the structural characteristics of book texts, we designed a prompt template with hierarchical constraints. By inputting the interpreted content generated by the agent into this customized prompt template, it effectively identifies the hierarchical knowledge units within the book's content and outputs standardized JSON data, which is then further visualized by the front end. Specifically, the LLMs need to perform the following operations: (1) root node identification, extracting the core themes of the book as the root node of the mind map; (2) multi-level node division, where the first-level nodes represent main arguments, second-level nodes represent key subtopics and sub-arguments, and third-level nodes represent specific evidence, methods, and explanations. By integrating the "auditory + visual" dual-service model, this approach not only enhances the depth of user understanding of the book’s content but also significantly improves the intuitiveness and interactivity of information transmission, thereby providing users with a more efficient learning and reading experience.

% \section{Evaluation Website}

% \begin{figure*}[h!]
%     \centering
%     \includegraphics[width=\linewidth]{latex/figures/evaluation_1.png}
%     \caption{Screenshots of system interface.}
%     \label{fig:evaluation_1}
% \end{figure*}
% \begin{figure*}[h!]
%     \centering
%     \includegraphics[width=\linewidth]{latex/figures/evaluation_2.png}
%     \caption{Screenshots of system interface.}
%     \label{fig:evaluation_2}
% \end{figure*}

% 
% \section{}

\end{document}